\documentclass{article}

\usepackage{arxiv}

\usepackage[utf8]{inputenc} 
\usepackage[T1]{fontenc}    
\usepackage{hyperref}       
\usepackage{url}            
\usepackage{booktabs}       
\usepackage{amsfonts}       
\usepackage{nicefrac}       
\usepackage{microtype}      
\usepackage{lipsum}		
\usepackage{graphicx}
\usepackage{natbib}
\usepackage{doi}

\title{Multi-atlas Ensemble Graph Neural Network
Model For Major Depressive Disorder
Detection Using Functional MRI Data}

\author{ \href{https://orcid.org/0000-0000-0000-0000}{\includegraphics[scale=0.06]{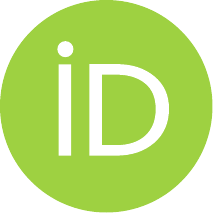}\hspace{1mm}Nojod M. Alotaibi*}\\
        Department of Computer Science, \\
        Faculty of Computing and Information Technology,\\
        King Abdulaziz University, Jeddah, Saudi Arabia\\
        \texttt{nalotaibi0351@stu.kau.edu.sa}\\
	\And
	\href{https://orcid.org/0000-0000-0000-0000}{\includegraphics[scale=0.06]{orcid.pdf}\hspace{1mm}Areej M. Alhothali*}\\
        Department of Computer Science, \\
        Faculty of Computing and Information Technology,\\
        King Abdulaziz University, Jeddah, Saudi Arabia\\
        \texttt{aalhothali@kau.edu.sa}\\
    \And
	\href{https://orcid.org/0000-0000-0000-0000}{\includegraphics[scale=0.06]{orcid.pdf}\hspace{1mm} Manar S. Ali} \\
	Department of Computer Science, \\
        Faculty of Computing and Information Technology,\\
        King Abdulaziz University, Jeddah, Saudi Arabia\\
}

\renewcommand{\shorttitle}{\textit Multi-atlas Ensembel GNN Model}

\hypersetup{
pdftitle={Multi-atlas Ensemble Graph Neural Network Model For Major Depressive Disorder Detection Using Functional MRI Data},
pdfsubject={q-bio.NC, q-bio.QM},
pdfauthor={Nojod M. Alotaibi*, Areej M. Alhothali*, and Manar S. Ali},
pdfkeywords={Major depressive disorder, rs-fMRI, brain functional connectivity network, deep learning, graph neural network, data oversampling, ensemble model},
}
\usepackage{multirow}
\usepackage{multicol}
\usepackage[table]{xcolor}
\begin{document}
\maketitle

\begin{abstract}
	Major depressive disorder (MDD) is one of the most common mental disorders, with significant impacts on many daily activities and quality of life. It stands as one of the most common mental disorders globally and ranks as the second leading cause of disability. The current diagnostic approach for MDD primarily relies on clinical observations and patient-reported symptoms, overlooking the diverse underlying causes and pathophysiological factors contributing to depression. Therefore, scientific researchers and clinicians must gain a deeper understanding of the pathophysiological mechanisms involved in MDD. There is growing evidence in neuroscience that depression is a brain network disorder, and the use of neuroimaging, such as magnetic resonance imaging (MRI), plays a significant role in identifying and treating MDD. Rest-state functional MRI (rs-fMRI) is among the most popular neuroimaging techniques used to study MDD. Deep learning techniques have been widely applied to neuroimaging data to help with early mental health disorder detection. Recent years have seen a rise in interest in graph neural networks (GNNs), which are deep neural architectures specifically designed to handle graph-structured data like rs-fMRI. This research aimed to develop an ensemble-based GNN model capable of detecting discriminative features from rs-fMRI images for the purpose of diagnosing MDD. Specifically, we constructed an ensemble model by combining features from multiple brain region segmentation atlases to capture brain complexity and detect distinct features more accurately than single atlas-based models. Further, the effectiveness of our model is demonstrated by assessing its performance on a large multi-site MDD dataset. The best performing model among all folds achieved an accuracy of 75.80\%, a sensitivity of 88.89\%, a specificity of 61.84\%, a precision of 71.29\%, and an F1-score of 79.12\%. 
\end{abstract}


\section{Introduction}
Mental disorders are considered a major cause of disability and are correlated with a higher risk of early death~\citep{freeman2022world}. There is a strong correlation between mental disorders, suicide, and other chronic diseases, such as cardiovascular disease, diabetes, and HIV/AIDS~\citep{freeman2022world}. Major depressive disorder (MDD) is a serious mental disorder that leads to severe disruptions in the ways in which an individual expresses emotions, rationalizes, and engages in social interactions~\citep{zhuo2019rise}. It is among the most prevalent mental disorders. According to the Global Burden of Diseases, Injuries and Risk Factors Study 2020 (GBD 2020), MDD prevalence increased by $27.6\%$, since the COVID-19 outbreak~\citep{santomauro2021global}. Before the pandemic, an estimated $193$ million people suffered from MDD. After the pandemic, the number of individuals experiencing MDD jumped to 246 million ~\citep{santomauro2021global}. Furthermore, MDD was the second primary cause of disability in the world, contributing significantly to the global disease burden in $2019$~\citep{gbd2022global}. 

At present, diagnosing MDD is primarily dependent on criteria provided by the Diagnostic and Statistical Manual of Mental Disorders (DSM) along with self-reported symptoms obtained through clinical questionnaires and clinical interviews~\citep{zhuo2019rise}. As MDD shares many symptoms with other mental disorders, distinguishing them can be challenging, requiring the expertise of highly trained psychiatrists. Thus, such methods of diagnosis are time consuming and unable to detect MDD in its earliest stages. Moreover, the lack of recognized neurological biomarkers for MDD makes the diagnostic and prognostic process difficult~\citep{zhuo2019rise}. Actually, there is evidence from neuroimaging studies that it is closely related to abnormality functionally and structurally in certain regions of the brain~\citep{pilmeyer2022functional}. In the last decade, neuroimaging has gained considerable prominence due to advances in computing technology, which have enabled researchers to obtain a deeper comprehension of the brain’s mechanisms~\citep{tulay2019multimodal}. Resting-state functional magnetic resonance imaging (rs-fMRI) has become a widely used neuroimaging method for analyzing MDD as it offers insights into the brain’s functional connectivities by detecting dynamic changes in blood oxygenation-dependent signals (BOLDs) of patients~\citep{pilmeyer2022functional}. Several studies have established that functional connectivity networks (FCNs) are reliable data sources for diagnosing MDD as they represent correlations between different brain regions of interest (ROIs)~\citep{smitha2017resting}. The rs-fMRI findings indicated that patients with MDD had abnormal functional connectivity related to the central executive network (CEN), the default mode network (DMN), and the salience network (SN)~\citep{dai2019brain}. Several studies based on rs-fMRI also found that MDD patients have abnormal brain function in several cortical and subcortical structures, including the prefrontal cortex, amygdala, insula, hippocampus, and precuneus~\citep{dai2019brain}.

Currently, most neuroimaging-based MDD studies are primarily focused on exploring functional imaging biomarkers, by utilizing different machine learning (ML) and deep learning (DL) approaches to analyze complex medical images and assist physicians in making an accurate diagnosis. Among these approaches are support vector machines (SVM), Random Forest (RF), 3D convolutional neural networks (CNNs), BrainNetCNN, and ResNet50. Graph neural networks (GNNs) are recently emerging as a new type of DL technique capable of analyzing graph-structured data, such as 4D rs-fMRI, which are difficult to model directly using conventional DL methods created for Euclidean data such as images and texts~\citep{ahmedt2021graph}. Several studies have been conducted to detect MDD using rs-fMRI, including,~\citep{yao2020temporal}, \citep{qin2022using}, and \citep{noman2024graph}. Even though the studies mentioned above have proven successful in diagnosing MDD, they only used a single brain region segmentation template (atlas). Nevertheless, the heterogeneity and complexity of brain regions cannot be fully captured using a single brain atlas~\citep{liu2018mmm}. To tackle this issue, several studies have been developed that use multi-atlas fusion to detect mental disorders, including \citep{liu2016relationship}, \citep{LEI2020101632}, \citep{chu2022multi}, \citep{xia2023depressiongraph}, and \citep{lee2024spectral}. However, most existing studies have concatenated or averaged two or three atlases to provide complementary information about the brain. They also trained their models using a relatively small sample size. In fact, rs-fMRI datasets have a smaller sample size than traditional datasets used in ML since they are more challenging to obtain and analyze. As a result, applying DL methods to rs-fMRI datasets may lead to model underfitting or overfitting~\citep{liu2023spatial}. Consequently, large-scale datasets are more suitable for training DL models, which will enhance the models' classification performance and make them more dependable and generalizable~\citep{alzubaidi2023survey}. To address this issue, many studies have suggested oversampling techniques to generate new samples from the rs-fMRI dataset, including~\citep{venkatapathy2023ensemble} and~\citep{liu2023spatial}.  

This research is designed to develop an ensemble-based GNN model able to categorize rs-fMRI data into MDD and healthy control (HC) subjects. Our model consists of four GNNs, each of which is trained using FCNs derived from a different brain atlas. Models based on multiple atlases can capture the complexity of the human brain structure and detect more discriminatory features than models based solely on an atlas. Moreover, the Synthetic Minority Over-sampling (SMOTE) technique is applied to produce a wide variety of data suitable for training our ensemble model. To assess the effectiveness of the proposed model, a comprehensive evaluation is conducted on a substantial multi-site MDD dataset, allowing for generalizable and reliable classification performance for MDD.

The rest of the paper is organized as follows. In Section 2, we provide a brief overview of the most relevant studies. Our proposed method is introduced in Section 3 by providing an overview of the materials and methods used to develop it. Section 4 presents experimental settings and evaluation metrics, followed by a discussion of the classification results. Finally, the conclusion of this paper is summarized in Section 5.

\section{Related Works}
Recently, a number of studies have been conducted on the detection of MDD based on rs-fMRI features. Among these studies, Yao et al.~\citep{yao2020temporal} introduced a temporal-adaptive graph convolutional network (TAGCN) model for capturing both dynamic and static information about functional connectivity (FC) patterns towards the diagnosis of MDD. The rs-fMRI time-series signals were first extracted from each ROI and divided into multiple overlapped blocks using fixed-size sliding windows. An adaptive graph convolutional layer was then applied to produce dynamic connectivity matrices for each block. Afterwards, convolution operations were performed on each ROI along various blocks to obtain the temporal dynamics of the entire time series. Finally, the classification of MDD was accomplished through a fully connected layer and a softmax function. The TAGCN method performed better than the other methods and achieved accuracy of $73.8\%$. In Qin et al. \citep{qin2022using}, a model was developed to support early and precise diagnosis of MDD using rs-fMRI data by applying GCN to a large multi-site MDD dataset. A whole-brain functional network was used to train the GCN model to recognize MDD patients from HC. Further, the authors conducted a subgroup analysis to distinguish between patients with first-episode drug-naive (FEDN) and HC, recurrent and HC, and recurrent MDD and FEDN. Results showed that GCN had an accuracy of 81.5\%, which was higher than that of other competing classifiers. Noman et al.~\citep{noman2024graph} constructed a graph DL framework for classifying fMRI-derived brain networks in MDD using non-Euclidean information about graph structure. Based on GCNs, a novel graph autoencoder (GAE) architecture was designed to encode the topological structure and node content into low-dimensional latent representations and train a decoder for graph reconstruction. In addition, the authors developed a framework to learn graph embeddings in brain networks in an unsupervised manner. A deep fully connected neural network (FCNN) was then trained to recognize MDD from HC using the learned embeddings. It was found that GAE-FCNN significantly outperformed other existing methods, attaining the highest accuracy of $72.5\%$. 

In the study by Kong et al.~\citep{kong2021spatio}, a spatiotemporal GCN (STGCN) framework was developed to automatically diagnose MDD patients and predict their response to treatment based on FC. First, dynamic FCN were derived from rs-fMRI using a sliding temporal window method. Then, the spatial graph attention convolution module (SGAC) was created to enhance feature learning, while prior pooling was implemented to reduce feature dimensions. The temporal fusion module was designed for capturing dynamic FCN features among adjacent sliding windows together with the SGAC module. The STGCN achieved the highest diagnostic accuracy of 84.14\% and 83.93\% for the two datasets, respectively. The STGCN also achieved an accuracy of $89.63\%$ in predicting treatment response. A combination of 3D-CNN and cross-sample entropy (CSE) analysis was used by Lin et al.~\citep{lin2023automatic} to discriminate late-life depression patients from their non-depressed counterparts and to estimate depression severity level through brain fMRI data. First, the depression diagnosis network (DDN) was developed to distinguish MDD patients from HC. The depression severity prediction network (DSPN) was then used to predict symptoms severity for those who had been determined to be depressed. In addition, the Hamilton depression (HAM-D) scale was predicted based on the severity level of a depressed subject using a regression network. It was found that the proposed CSE-based CNN model was capable of attaining an accuracy rate of over $85\%$ for diagnosis. Kong et al.~\citep{kong2022multi} proposed a novel multi-stage graph fusion network (MSGFN) for the diagnosis of MDD. There were three main parts to this framework: FC calculation, multi-stage graph construction, and graph convolutional fusion. The purpose of FC calculation was to assess the interactions between white matter (WM) and gray matter (GM). A deep subspace model was utilized to derive multi-stage features from FC features in the multi-stage graph construction, and self-expression constraints were applied to create graphs at each stage. The graph convolutional fusion component was created to combine features and graphs from all stages. A comparison of MSGFN with other ML/DL models revealed that it achieved the highest accuracy of $70.91\%$.

Another group of studies utilized oversampling techniques to improve classification performance. Among these studies, Venkatapathy et al.~\citep{venkatapathy2023ensemble} proposed an ensemble model that combines three GNN approaches: GraphSAGE, GAT, and GCN for classification of MDD and HC and to analyze subgroups between patients with REC and FEDN. Moreover, they performed random oversampling by copying data from minority classes, and random undersampling by selecting data from majority classes in order to achieve a balanced sample size. In the effort to classify MDD from HC, upsampling produced $71.8\%$ accuracy, while downsampling yielded $70.4\%$ accuracy. With  upsampled REC and HC data, the model yielded $91.6\%$ accuracy; in contrast, with downsampled data, the model produced 68.78\% accuracy. Ultimately, the model produced accuracy of $77.78\%$ after upsampling REC with FEDN, and accuracy of $71.96\%$ after downsampling. Liu et al~\citep{liu2023spatial} created a spatial-temporal data-augmentation-based classification scheme (STDAC) that can fuse spatial-temporal information, enhance classification performance, and expand the sample size. A spatial data augmentation (SDA) module was built using KNN-like techniques, and a temporal data augmentation (TDA) module was constructed using discontinuous time series from the original data time period. Finally, the features from the aforementioned two modules were combined using a tensor fusion technique. Extensive experiments were conducted on the Alzheimer's Disease Neuroimaging (ADNI) and REST-meta-MDD datasets in order to assess the efficacy of the proposed model. Additionally, the STDAC model was trained using AAL atlas with some traditional ML classifiers, such as Random Forest (RF) and Neural Network (NN). It turned out that the proposed STDAC model with NN yielded the best accuracy for the two datasets, at $84.170\%$ and $63.406\%$, respectively.

The last group of studies focused on the fusion of multiple atlases. Xia et al. \citep{xia2023depressiongraph} introduced a two-channel graph neural network (DepressionGraph) model to efficiently aggregate the graph information from the two channels based on the node number and the node feature number. In addition, a transform-encoder architecture was employed to extract relevant information from time-series FCN. Using a majority voting method, the prediction results of models trained with three different atlases were integrated to improve the MDD prediction model. DepressionGraph demonstrated superior performance when compared to previous studies, with an accuracy of 71.48\%. In Lee et al. \citep{lee2024spectral}, an innovative multi-atlas fusion technique using GCN for MDD diagnosis was developed that integrates both early and late fusion approaches. During the early fusion phase, ROI signals are combined from each individual atlas. Next, a holistic FCN is created by calculating FCs between all ROI signals derived from three atlases. In the late fusion phase, a soft voting ensemble method was employed to combine feature vectors derived from FCs extracted from each atlas and the holistic FCN. It was found that the proposed model based on the GCN achieved the highest performance with an accuracy of 68.88±2.55\%.

However, a majority of depression detection studies used single-site datasets with small sample sizes, which limits generalizability and causes significant variations in model's performance. These approaches are often based upon a single atlas, which present considerable limitations, including an incomplete representation of brain structures and a potential bias imposed by the specific atlas. To address these challenges, we proposed a multi-atlas ensemble model leveraging a graph neural network (GNN) for categorizing rs-fMRI data into MDD and healthy controls (HC). This model improves representational power by incorporating diverse brain structures from multiple atlases, thus enhancing robustness and sensitivity to subtle connectivity changes. Besides that, we used the Synthetic Minority Over-sampling Technique (SMOTE) to produce a variety of training data allowing us to improve generalizability and reduce overfitting. 

\section{Materials and Methods}
This section provides a detailed description of the materials and methodologies employed in this study to differentiate MDD patients from HCs using rs‐fMRI time series. 
\subsection{REST-meta-MDD Dataset}
In this research, we utilize a dataset from a public open-access data repository called the REST-meta-MDD consortium, which is the largest public MDD dataset to our knowledge~\citep{chen2022direct}. The dataset consists of $2428$ subjects from $25$ sites, including $1300$ patients with MDD ($826$ are females and $474$ are males) and $1128$ HC~\citep {chen2022direct}~\citep{yan2019reduced}. At each site, phenotypic data such as age, sex, episode status (recurrent or first episode), medication status, illness duration, and the 17-item Hamilton Depression Rating Scale (HAMD) were gathered. There are also two types of imaging data available, T1-weighted sMRI and rs-fMRI~\citep{chen2022direct}. Participants in REST-meta-MDD were required to provide written informed consent before participating, and the local Institutional Review Board approved the collection of data at each site \citep {chen2022direct}~\citep{yan2019reduced}.
\subsubsection{Data Preprocessing}
At each local site, resting-state functional MRI and three-dimensional structural T1-weighted MRI images were acquired for all participants. The Data Processing Assistant (DPARSF) toolbox was used to perform a unified image preprocessing protocol. Among the preprocessing steps were discarding the first ten volumes to ensure magnetization equilibrium, correcting slice timing and head motion, normalizing to the Montreal Neurological Institute (MNI) template, smoothing, detrending and band-pass filtering, excluding sites with fewer than ten subjects, and regressing covariates \citep{yan2019reduced}. Further, our analysis of the time series of brain regions using targeted atlases revealed that some brain regions of some subjects were missing signals, so these subjects were excluded. Finally, $1563$ participants from $16$ sites were included in our study, of which $810$ MDD patients and $753$ HC. At the end of this process, the data for each subject was transformed into a matrix of size $T\times N$, where $T$ denotes the number of time points and $N$ denotes the number of brain ROIs. In this case, we used $140$ time points between each node. Detailed demographic information about the subjects included in our study is provided in Table \ref{DemoInfo}.

\begin{table}[h]
\centering
\caption{Demographic Information of the 1563 Study Subjects.}
\label{DemoInfo}
    \renewcommand{\arraystretch}{1.5}
    \begin{tabular}{|c|c|c|c|c|c|} \hline 
         \multirow{2}{*}{\textbf{Group}} & 
         \multirow{2}{*}{\textbf{Number of subjects}}& 
         \multirow{2}{*}{\textbf{Male}}& 
         \multirow{2}{*}{\textbf{Female}} &
         \multicolumn{2}{c|}{\textbf{Average $\pm$ Standard deviation (\%)} }\\ 
         \cline{5-6}
         &&&&
         \textbf{Age} & \textbf{Education}\\
         \hline 
         \
         \begin{tabular}{c} MDD  \end{tabular}& 
         \begin{tabular}{c}810\end{tabular}&  \begin{tabular}{c}293\end{tabular}&  \begin{tabular}{c}517\end{tabular}& \begin{tabular}{c}34.39 $\pm$ 11.55\end{tabular}& \begin{tabular}{c}11.96 $\pm$ 3.37\end{tabular}\\ \hline
         \begin{tabular}{c}HC  \end{tabular}&  \begin{tabular}{c}753\end{tabular}&  \begin{tabular}{c}307\end{tabular}&  \begin{tabular}{c}446\end{tabular}& \begin{tabular}{c}34.61 $\pm$ 13.17\end{tabular}& \begin{tabular}{c}13.57 $\pm$ 3.42\end{tabular}\\ \hline
    \end{tabular}
\end{table}
To overcome the challenges related to multi-site data, we performed ComBat harmonization to mitigate site-varying effects (also called batch effects), while keeping critical biological covariates such as sex and age \citep{el2023harmonization}. ComBat relies on a multivariate linear mixed-effects regression framework, which was developed for batch effect corrections in genomic studies \citep{johnson2007adjusting} \citep{el2023harmonization}. This model utilizes empirical Bayes to estimate both biological and non-biological effects and effectively remove the estimated additive and multiplicative site effects \citep{el2023harmonization}. As our data consists of time series features represented as $140 \times N$ matrices, where $140$ is the number of time points and $N$ is the number of brain ROIs, our ComBat harmonization model can be expressed mathematically as follows:
\begin{equation} \label{Combat}
Y_{ij} = \alpha + X_{ij} \beta + \gamma_i + \delta_i \epsilon_{ij}
\end{equation}

In this case, $Y_{ij}$ denotes the time series value for subject $j$ at site $i$, $\alpha$ is the overall mean, $X_{ij}$ is a design matrix for the biological covariates (i.e. sex and age), $\beta$ is the regression coefficients for covariates, $\gamma_i$ and $\delta_i$ are the site-specific effects (additive and multiplicative effects), and $\epsilon_{ij}$ represents the residual error. 
\subsection{Proposed Model}
\begin{figure}[htp!]
    \centering
    \graphicspath{{FigureFolder/}}
    \includegraphics[width=0.96\textwidth]{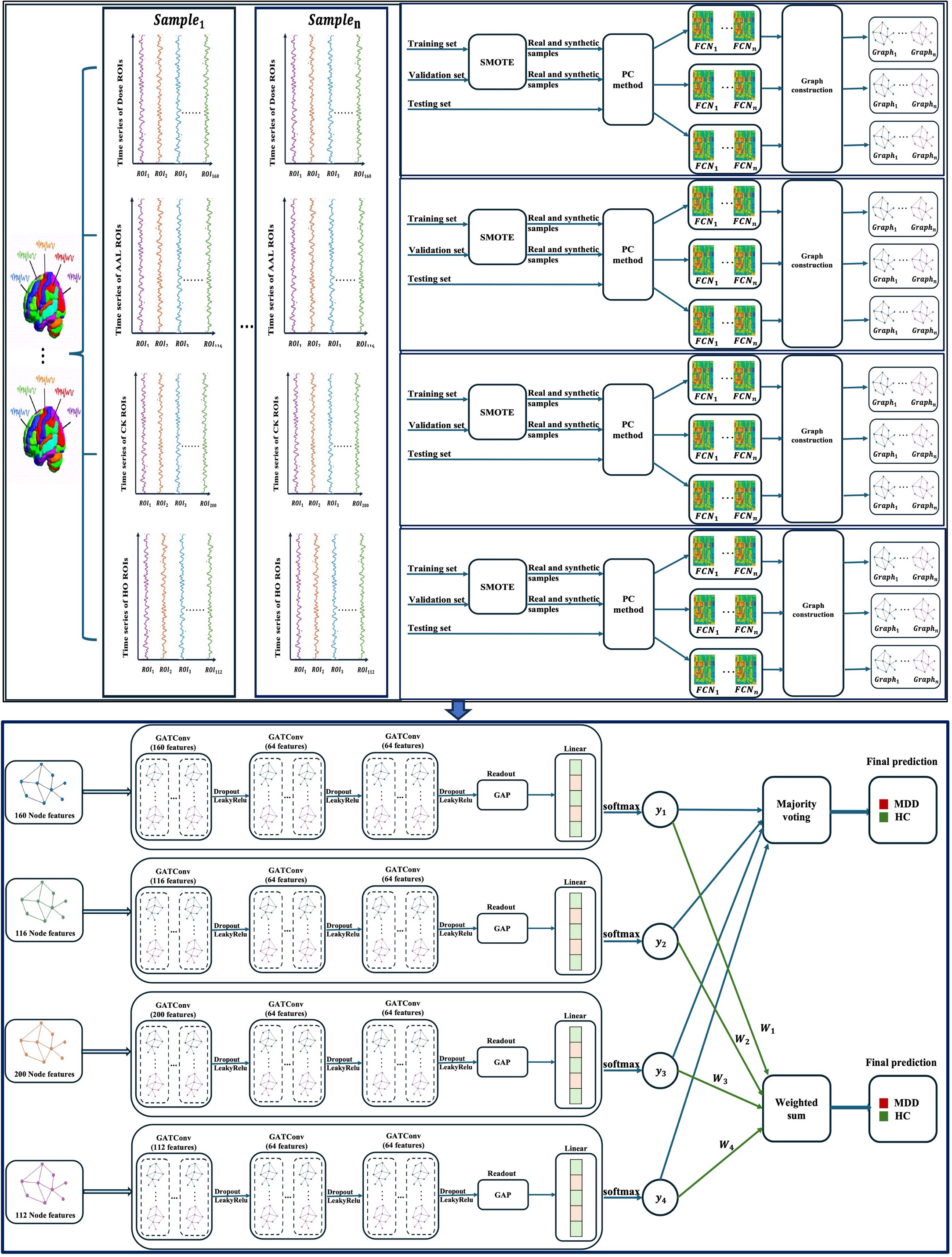}
    \caption{An overview of the multi-atlas ensemble GAT model. Abbreviations: GAT, graph attention network; Dose, Dosenbach’s atlas; AAL, Automated Anatomical Labeling atlas; CK, Craddock’s clustering atlas; HO, Harvard–Oxford atlas; ROI, region of interest; FCN, functional connectivity network; GAP, global average pooling; MDD, major depressive disorder; HC, healthy control; n, the number of samples.}
    \label{fig:Subfigure 1}
\end{figure}

This paper presents a multi-atlas ensemble GNN model for categorizing rs-fMRI data into MDD and HC subjects, as shown in Figure~\ref{fig:Subfigure 1}. As part of our study, we utilized four popular brain segmentation atlases, including Dosenbach's $160$ functional ROIs (Dose), Automated Anatomical Labeling (AAL) atlas, Craddock’s clustering $200$ ROIs (CK), and Harvard–Oxford (HO) atlas. The use of a variety of brain atlases contributes to improving the accuracy of diagnosis of brain disorders \citep{xia2023depressiongraph}. We first extracted time series for each atlas from the rs-fMRI data. Secondly, the Synthetic Minority Over-sampling (SMOTE) technique is adopted to generate a wide variety of data suitable for training our ensemble model. Thirdly, we generated FCNs for each atlas separately and represented them as graphs. Lastly, four homogeneous GNN models are constructed to complement our ensemble model, each having been trained with graphs derived from a different brain atlas. We used the GAT models as the base models for our multi-atlas ensemble model. 

\subsubsection{Graph Attention Network}
A graph attention network (GAT) is an innovative neural network architecture that employs masked self-attentional layers to address the limitations of prior approaches utilizing graph convolutions or their approximations~\citep{velivckovic2017graph}. The GAT model implements a self-attention strategy in which the representation of each node is computed by repeatedly attending to its neighbors. Initially, a graph attention layer receives a set of node features as input, $h=\{\vec{h}_{1}, \vec{h}_{2},..., \vec{h}_{N}\}$, where $N$ represents the number of nodes. This layer generate new node features, whose dimensions may differ from those of the input features. Next, self-attention is applied to derive attention coefficients $e_{ij}$ between each pair of nodes, as indicated in Equation~\ref{GATAttention} ~\citep{velivckovic2017graph}. 
\begin{equation} \label{GATAttention}
e_{ij}=a(W \vec{h}_{i}, W \vec{h}_{j})
\end{equation}

In this formula, $e_{ij}$ reflects the importance of node $j$'s attributes to node $i$, $a$ refers to the attention mechanism, and $W$ is a linear weight matrix shared by all nodes. Typically, attention coefficients are determined based on first-order neighbors. Further, a softmax function is used to normalize coefficients across all choices of $j$, enabling them to be compared across nodes. This process is illustrated in Equation~\ref{softmax} and Equation~\ref{leaky} \citep{velivckovic2017graph}. In Equation \ref{softmax}, the node $j$ $\in$ $N_{i}$, where $N_{i}$ refers to some neighborhood of node $i$.
\begin{equation} \label{softmax}
\alpha_{ij}=softmax_{j}(e_{ij})=\frac{exp(e_{ij})}{\sum_{k \in N_{i}} exp(e_{ik})}
\end{equation}
 
\begin{equation} \label{leaky}
\alpha_{ij}=\frac{exp(LeakyReLU(\vec{a}^T[W \vec{h}_{i} \parallel W \vec{h}_{j}]))}{\sum_{k \in N_{i}} exp(LeakyReLU(\vec{a}^T[W \vec{h}_{i}\parallel W \vec{h}_{k}]))}
\end{equation}

Where $\alpha_{ij}$ is normalized attention coefficient, $\vec{a}$ is a weight vector, LeakyReLU is a nonlinear function, ${.}^{T}$ refers to a transposition, and $\parallel$ is the concatenation operation. Lastly, these normalized attention coefficients are utilized to calculate a linear combination of nodes' features, resulting in a new embedding for each node (Equation \ref{Embedding}) \citep{velivckovic2017graph}.
\begin{equation} \label{Embedding}
\vec{h}_{i}^{'}=\sigma(\sum_{j \in N_{i}} \alpha_{ij} W \vec{h}_{j})
\end{equation}

Where $\sigma$ represents a nonlinear function. Assuming that $K$ distinct attention heads have been used to compute the node embedding, then the results should be concatenated (Equation \ref{multihead1}) or averaged (Equation \ref{multihead2}) as follows  \citep{velivckovic2017graph}:
\begin{equation} \label{multihead1}
\vec{h}_{i}^{'}=\parallel_{k=1}^{K} \sigma(\sum_{j \in N_{i}} \alpha_{ij}^{k} W^{k} \vec{h}_{j})
\end{equation}
\begin{equation} \label{multihead2}
\vec{h}_{i}^{'}= \sigma(\frac{1}{K} \sum_{k=1}^{K}\sum_{j \in N_{i}} \alpha_{ij}^{k} W^{k} \vec{h}_{j})
\end{equation}

Where $\parallel$ is concatenation, $\alpha_{ij}^{k}$ are normalized attention coefficients calculated by the $k$-th attention mechanism, $\sigma$ is a nonlinear function, and $W$ is a linear weight matrix. 

\subsubsection{Data Oversampling}
In general, deep learning models have a large number of parameters that must be optimized during the training process. Therefore, training deep learning models on large datasets increases their reliability, generalizability, and their classification performance~\citep{alzubaidi2023survey}. As rs-fMRI datasets are difficult to acquire and collect, their sample size is smaller than classic datasets used for machine learning. Thus, there is a risk of underfitting or overfitting the model when analyzing FBN data~\citep{liu2023spatial}. To overcome this issue, many studies have proposed oversampling techniques that generate new data samples from the original dataset. In real-world applications, the synthetic minority oversampling (SMOTE) technique is one of the most widely used oversampling methods due to its simplicity, high performance, and computational efficiency~\citep{mansourifar2020deep}. In SMOTE, the oversampling is carried out by creating one or more synthetic samples for each training point belonging to the minority class. Specifically, these synthetic samples are obtained in feature space by applying linear interpolation between minority class samples and their K nearest neighbors \citep{chawla2002smote}. As a first step, the k-nearest neighbors of sample are computed, and one of them is chosen randomly. Next, the difference between the feature vector of the real sample $X_{0}$ and its nearest neighbor $X$ is computed. This difference is then multiplied by a random number between $0$ and $1$, known as the gap, and then added to the real sample's feature vector to produce a synthetic feature vector $Z$~\citep{chawla2002smote}. These steps are repeated for each sample based on how many synthetic samples are needed. The Equation~\ref{eqSMOTE} used for constructing synthetic samples is as follows~\citep{chawla2002smote}:

\begin{equation}\label{eqSMOTE}
Z=X_0+\left(X-X_0\right) \times gap
\end{equation}

In this study, SMOTE was used to oversample MDD and HC classes to double both training and validation sets. Although the dataset is nearly balanced regarding the number of individuals with MDD and HC, it is not balanced in terms of the samples per site. Consequently, SMOTE was applied to generate synthetic samples that reduce overfitting and balance data across sites. Firstly, we divided the original time series dataset into $80\%$ for training, 10\% for validation, and $10\%$ for testing. Next, each of the training and validation sets is divided into two sets, one containing MDD samples and the other containing HC samples. After that, oversampling was performed using SMOTE on each MDD and HC samples independently. Our current implementation uses three nearest neighbors to interpolate new synthetic samples. This resulted in $2502$ training samples: $1298$ MDD samples and $1204$ HC samples, and $310$ validation samples: 160 MDD samples and $150$ HC samples. We oversampled only the training and validation sets in order to ensure that only real data were included in the testing set. Consequently, there are 157 real testing samples: 81 MDD samples and 76 HC samples.

\subsubsection{The construction of the brain network}

In graph theory, the graph structure is expressed as $G = \{V, E, A\}$, where $V$ represents a set of vertices or nodes, $E$ represents a set of edges between these nodes, and $A$ is an adjacency matrix~\citep{lima2022comprehensive}. An adjacency matrix illustrates the relationships between each pair of nodes in a graph, in which the connection between a given pair of nodes is determined by the entry of $A$ in the $i$-th row and $j$-th column of the matrix, and is denoted as $A_{ij}$. In this matrix, element $A_{ij}$  is 1 if there is an edge from node $i$ to node $j$ else $A_{ij}$ is 0. Furthermore, a graph is considered undirected if none of its edges have a direction, whereas it is considered directed if all its edges have a direction. A graph can also have weighted edges, in which the adjacency matrix entries are arbitrary real-values instead of $0$ and $1$~\citep{wu2022graph}. Feature vectors can also be attached to nodes in a graph. In this case, $x_{i}$ indicates the feature vector for the node $v_{i}$, and $X_V$ = $[x_1, x_2,..., x_n]$ represents the set of feature vectors for all the nodes in the graph. Similarly, graph edges may be associated with edge feature vectors $x(i , j)$.

Based on rs-fMRI time series, we first constructed a functional connectivity network/matrix (FCN) for representing each subject by computing the correlation between each pair of ROIs using Pearson correlation coefficients (PC), as shown in Equation~\ref{eqPC}~\citep{wang2022adaptive}. Fisher’s z-transformation was then applied to these matrices to standardize correlations across different sites \citep{vergara2018method} \citep{dai2024classification}. Second, an undirected graph was generated from each brain FCN, where graph nodes represent brain regions of interest (ROIs) and node features are determined by the rows in FCN. To establish edges, the k-nearest neighbor (KNN) algorithm was employed to connect each node with its neighbors. Further, the relationships between the nodes are represented by a weighted adjacency matrix $A$.
\begin{equation} \label{eqPC}
b_{i j}=\frac{\left(y_i-\bar{y}_i\right)\left(y_j-\bar{y}_j\right)}{\sqrt{\left(y_i-\bar{y}_i\right)\left(y_i-\bar{y}_i\right)} \sqrt{\left(y_j-\bar{y}_j\right)\left(y_j-\bar{y}_j\right)}}
\end{equation}

Where $b_{ij}\in [-1,1]$ is the correlation between the i-th and j-th ROIs, $y_{i}$ and $y_{j}$ correspond to time series points for ROIs $i$ and $j$, while $\bar{y}_{i}$ and $\bar{y}_{j}$ represent the means of the regional rs-fMRI singnals in ROIs $i$ and $j$, respectively.

\subsubsection{Ensemble Methods}
Ensemble learning is the process of combining several baseline models to construct a more powerful and generalizable model than its constituent models~\citep{ mohammed2023comprehensive}. A typical ensemble learning system relies on an aggregation function $G$ to combine a set of baseline classifiers $C_{1}$,$C_{2}$,…,$C_{n}$ to predict a single output. The aggregation function is can be derived using different techniques, such as majority voting, weighted voting, averaging, weighted averaging, sum, and weighted sum~\citep{ mohammed2023comprehensive}.  

In this regard, we developed a multi-atlas ensemble model consisting of four GAT models, each trained on graphs obtained from a different brain atlas. Following this, the GAT models make a set of predictions based on their respective test data, which are then combined to produce a final ensemble prediction. To accomplish this, majority voting and weighted sum methods were applied. In majority voting (or hard voting), each model $C_{i}$ in the ensemble separately predicts the class label for a given input, which is considered a vote. Afterwards, the ensemble determines the final label prediction $\widehat{y}$ by choosing the class label that takes the most votes from the individual models as shown in Equation~\ref{eqVote} \citep{ mohammed2023comprehensive}. In this Equation, the mode function refers to the most frequently occurring value within a set of values. 
\begin{equation} \label{eqVote}
\widehat{y}= mode[C_{1}(x), C_{2}(x),..., C_{n}(x)]
\end{equation}

Since we have an even number of models, ties may occur if more than one class receives the same score. In this case, we broke the tie by taking the prediction from the model that had the highest accuracy. For weighted sum, the output of each model is taken after applying the softmax function. This produces a vector for each input containing the probability of each class label. The probability values for each class label are then multiplied by the model weight. In this study, each model's weight is determined by its average accuracy based on the validation set, as indicated in Equation~\ref{eqWeight}.
\begin{equation}\label{eqWeight}
w_{j}=\frac{Acc_{j}}{\sum_{i=1}^n Acc_{i}}    
\end{equation}

Assume that $w_j$ is the weight of the model $j$, $n$ is the number of models, and $Acc_i$ is the accuracy of the $i^{th}$ model. After that, the weighted probabilities for each class are summed, and the class label with the highest sum probability is selected as the final ensemble prediction. A weighted sum method for a binary classification task is demonstrated in the following Equation~\ref{eqWeightSum}~\citep{mohammed2023comprehensive}.
\begin{equation}\label{eqWeightSum}
\widehat{y}=argmax_{i} \sum_{j=1}^n w_{j} \times p_{ij}
\end{equation}

In this case, $\widehat{y}$ represents the predicted class label, as determined by weighted sum, n represents the number of models, $w_j$ represents the weight of the model $j$, and $p_{ij}$ represents the probability of class $i$ in the model $j$.

\section{Results and Discussion}
Throughout this section, we first introduce the implementation details and the evaluation metrics for the classification task. Then, we present and discuss the prediction performance of the proposed model on a large-scale fMRI dataset obtained from the REST-meta-MDD project.
\subsection{Implementation details and Evaluation Metrics}
In this study, we developed an ensemble-based GNN model for identifying MDD and HC subjects based on the rs-fMRI data. This model is comprised of four GAT models each trained on a different atlas. For this purpose, several brain segmentation atlases were employed, including Dosenbach's 160 functional ROIs (Dose), Automated Anatomical Labeling (AAL), Craddock's clustering atlas (CK), and Harvard-Oxford (HO). A GAT model is composed of three stacked layers of GATConv with Leaky Rectified Linear Unit (LeakyReLU) activation functions, which are used to learn the representation of each node. Each GATConv layer is composed of $64$ hidden units. The number of attention heads used in this study is $4$. Before each GATConv layer, the dropout technique is performed to prevent overfitting. Further, global average pooling (GAP) is applied to generate a representation of the whole graph by averaging the final embeddings of each node, as indicated in Equation \ref{GAPEq} \citep{cui2022braingb}. In Equation \ref{GAPEq}, $h_{G}$ refers to the representation of a single graph $G$, $N$ is the number of nodes in $G$, and $h_i$ is the embedding of the $i$-th node in $G$. For graph classification, a fully connected (FC) layer is added and activated by a softmax function to convert output scalars into the predictive probability of each class. 
\begin{equation}\label{GAPEq}
    h_{G} = \frac{1}{N} \sum_{i=1}^{N} {h}_i
\end{equation}

During the training stage, the cross-entropy loss function is used to optimize the parameters of the model The loss function is regularized using the $L2$ regularization technique known as weight decay. Additionally, Adaptive Moment Estimation (Adam) is utilized as an optimization function to rapidly update all weights based on a constant learning rate alpha. We employed a grid search strategy to identify the optimal values of the hyperparameters to enhance the performance of the model. The final configuration included a batch size of $16$, learning rate of $0.001$, a weight decay value of  $5 \times 10^{-4}$, and dropout rate of $0.5$.

Furthermore, we evaluated the proposed model using the REST-meta-MDD dataset based on a stratified 10-fold cross-validation method. The stratified cross-validation method allows us to keep the proportion of samples from each class equal across all folds. Using this method, a dataset is divided into $10$ non-overlapping subsets. During each iteration, one subset is selected as a test set for model evaluation, while the remaining nine subsets are used as a training set. Afterwards, the performance of the proposed model is assessed using five widely used metrics, including accuracy (Acc), sensitivity (Sen), specificity (Spe), precision (Pre), and F1-score (F1). These metrics are calculated as follows \citep{SHARMA202231}:
\begin{equation} \label{eqAcc}
Acc =\frac{(TP + TN)}{(TP + TN + FP + FN)}
\end{equation}
\begin{equation} \label{eqSn}
Sen = \frac{TP}{(TP + FN)} 
\end{equation}
\begin{equation} \label{eqSp}
Spe = \frac{TN}{(FP + TN)} 
\end{equation}
\begin{equation} \label{eqP}
Pre = \frac{TP}{(TP + FP)}
\end{equation}
\begin{equation} \label{eqF1}
F1 = \frac{(2 \times Pre \times Sen)}{(Pre + Sen)}
\end{equation}

In this case, $TP$ denotes the correct classification of positive samples (MDD), $TN$ denotes the correct classification of negative samples (HC), $FP$ denotes the incorrect classification of negative samples as positive, and $FN$ denotes the incorrect negative classification.

\subsection{Classification Results}
This section examines the effectiveness of our suggested MDD classification method by analyzing four scenarios:
(1) A single-atlas GAT model with oversampling (SMOTE); (2) An multi-atlas ensemble GAT model with oversampling; (3) An examination of the statistical differences between the multi-atlas GAT model and each single-atlas GAT model; (4) A comparison of GAT models with and without SMOTE; (5) A comparison with single-site models; (6) A comparison with existing studies.

\subsubsection{Results of Single-Atlas GAT Models With SMOTE}
The classification results of a single GAT model applied to different brain atlas datasets using the SMOTE algorithm are presented in Table~\ref{SingleAt_SMOTE}. The results indicate that the CK-based model has superior performance compared to other models based on the Dose, AAL, and HO atlases in three metrics: an accuracy of 65.41±2.92\%, a sensitivity of 75.43±6.26\%, and an F1-score of 69.16±2.90\%. The HO-based model achieves the second highest performance with 64.33±1.07\% accuracy, 69.63±9.11\% sensitivity, and 66.59±2.78\% F1-score. Nevertheless, the specificity and precision of the HO-based model are 58.68±10.09\% and 64.65±2.70\%, which are significantly higher than that of other models based on a single atlas. The AAL-based model produces an accuracy of 63.95±4.01\%, a sensitivity of 70.99±7.27\%, a specificity of 56.45±11.88\%, a precision of 64.04±4.33\%, and an F1-score of 66.95±3.03\%. On the other hand, the Dose-based model achieves the lowest performance with 61.91±1.84\% accuracy, 71.23±8.50\% sensitivity, 51.97±8.87\% specificity, 61.47±2.29\% precision, and 65.66±3.32\% F1-score. 

Furthermore, the evaluation metrics for the fold that has the highest accuracy for each atlas-based model are displayed in Table~\ref{BestSingleAt}. In the CK-based model, the best model among all $10$ folds exhibits an accuracy of 70.06\%, a sensitivity of 86.42\%, a specificity of 52.63\%, a precision of 66.04\%, and an F1-score of 74.87\%. The best AAL-based model has 67.52\% accuracy, 76.54\% sensitivity, 57.89\% specificity, 65.96\% precision, and 70.86\% F1-score. While the best HO-based model has 66.24\% accuracy, 61.73\% sensitivity, 71.05\% specificity, 69.44\% precision, and 65.36\% F1-score. Using the Dose-based model, the best model achieves an accuracy of 65.61\%, a sensitivity of 65.43\%, a specificity of 65.79\%, a precision of 67.09\%, and an F1-score of 66.25\%.   
\begin{table}[h]
\caption{Average Classification Results and Standard Deviation of 10-Fold Cross-Validation for the Single-Atlas GAT Model with SMOTE.\\}
\label{SingleAt_SMOTE}

\resizebox{\columnwidth}{!}{%
\renewcommand{\arraystretch}{1.5}
\begin{tabular}{|c|c|c|c|c|c|}
\hline \multirow{2}{*}{ \textbf{Method}} & \multicolumn{5}{c|}{ \textbf{Average $\pm$ Standard deviation (\%)} } \\
\cline { 2 - 6 } & \textbf{Accuracy} & \textbf{Sensitivity} & \textbf{Specificity} & \textbf{Precision} & \textbf{F1-score} \\
\hline \begin{tabular}{c}Dose-based model\\ \end{tabular}& $61.91\pm1.84$ & $71.23\pm8.50$ & $51.97\pm8.87$ & $61.47\pm2.29$ & $65.66\pm3.32$ \\
\hline \begin{tabular}{c}AAL-based model\\ \end{tabular} & $63.95\pm4.01$ & $70.99\pm7.27$ & $56.45\pm11.88$ & $64.04\pm4.33$ & $66.95\pm3.03$ \\
\hline \begin{tabular}{c}CK-based model\\ \end{tabular} & $\mathbf{65.41\pm2.92}$ & $\mathbf{75.43\pm6.26}$ & $54.74\pm7.16$ & $64.15\pm2.91$ & $\mathbf{69.16\pm2.90}$ \\
\hline \begin{tabular}{c}HO-based model\\ \end{tabular} & $64.33\pm1.07$ & $69.63\pm9.11$ & $\mathbf{58.68\pm10.09}$ & $\mathbf{64.65\pm2.70}$ & $66.59\pm2.78$ \\
\hline
\end{tabular}
}
{\\\\The bolded values indicate the best performing results.}
\end{table}

\begin{table}[h]
\caption{Classification Results for the Fold with the Highest Accuracy from 10-Fold Cross-Validation for the Single-Atlas GAT Model with SMOTE.\\}
\label{BestSingleAt}
\resizebox{\columnwidth}{!}{%
\renewcommand{\arraystretch}{1.5}
\begin{tabular}{|c|c|c|c|c|c|}
\hline \textbf{Method} & \textbf{Accuracy (\%)} & \textbf{Sensitivity (\%)}& \textbf{Specificity (\%)} & \textbf{Precision (\%)} & \textbf{F1-score (\%)} \\
\hline \begin{tabular}{c}Dose-based model \end{tabular}& $65.61$ & $65.43$ & $65.79$ & $67.09$ & $66.25$ \\
\hline \begin{tabular}{c}AAL-based model\end{tabular} & $67.52$ & $76.54$ & $57.89$ & $65.96$ & $70.86$ \\
\hline \begin{tabular}{c}CK-based model\end{tabular} & $\mathbf{70.06}$ & $\mathbf{86.42}$ & $52.63$ & $66.04$ & $\mathbf{74.87}$ \\
\hline \begin{tabular}{c}HO-based model\end{tabular} & $66.24$ & $61.73$ & $\mathbf{71.05}$ & $\mathbf{69.44}$ & $65.36$ \\
\hline
\end{tabular}%
}
{\\\\The bolded values indicate the best performing results.}
\end{table}

\subsubsection{Results of Multi-Atlas GAT Ensemble Models With SMOTE}
We demonstrate in Table~\ref{ResultsEnsemble} the effectiveness of combining four brain atlas datasets for MDD prediction using three different ensemble methods. Based on the results, a majority voting ensemble outperforms other ensemble methods across all the metrics, including an accuracy of 69.49±3.54\%, a sensitivity of 78.02±6.55\%, a specificity of 60.39±5.41\%, a precision of 67.79±2.87\%, and an F1-score of 72.43±3.71\%. The ensemble model based on the sum method achieves the second best performance with 67.83±3.66\% accuracy, 77.78±6.51\% sensitivity, 57.24±8.71\% specificity, 66.24±3.77\% precision, and 71.34±3.28\% F1-score. Meanwhile, a weighted sum ensemble provides the lowest performance compared to other methods. It has an accuracy of 67.45±3.20\%, a sensitivity of 77.65±6.33\%, a specificity of 56.58±8.40\%, a precision of 65.84±3.45\%, and an F1-score of 71.06±2.94\%. In addition, the evaluation metrics for the fold with the highest accuracy for each ensemble model are presented in Table \ref{BestEnsemble}. In a majority voting ensemble, the best performing model among all 10 folds attains an accuracy of 75.80\%, a sensitivity of 88.89\%, a specificity of 61.84\%, a precision of 71.29\%, and an F1-score of 79.12\%.  Using the sum method, the best ensemble model produces 73.25\% accuracy, 85.19\% sensitivity, 60.53\% specificity, 69.70\% precision, and 76.67\% F1-score. On the other hand, the best performing model obtained by a weighted sum ensemble has 70.70\% accuracy, 82.72\% sensitivity, 57.89\% specificity, 67.68\% precision, and 74.44\% F1-score. 

\begin{figure}[htp!]
    \centering
    \graphicspath{{FigureFolder/}}
    \includegraphics[width=0.99\textwidth]{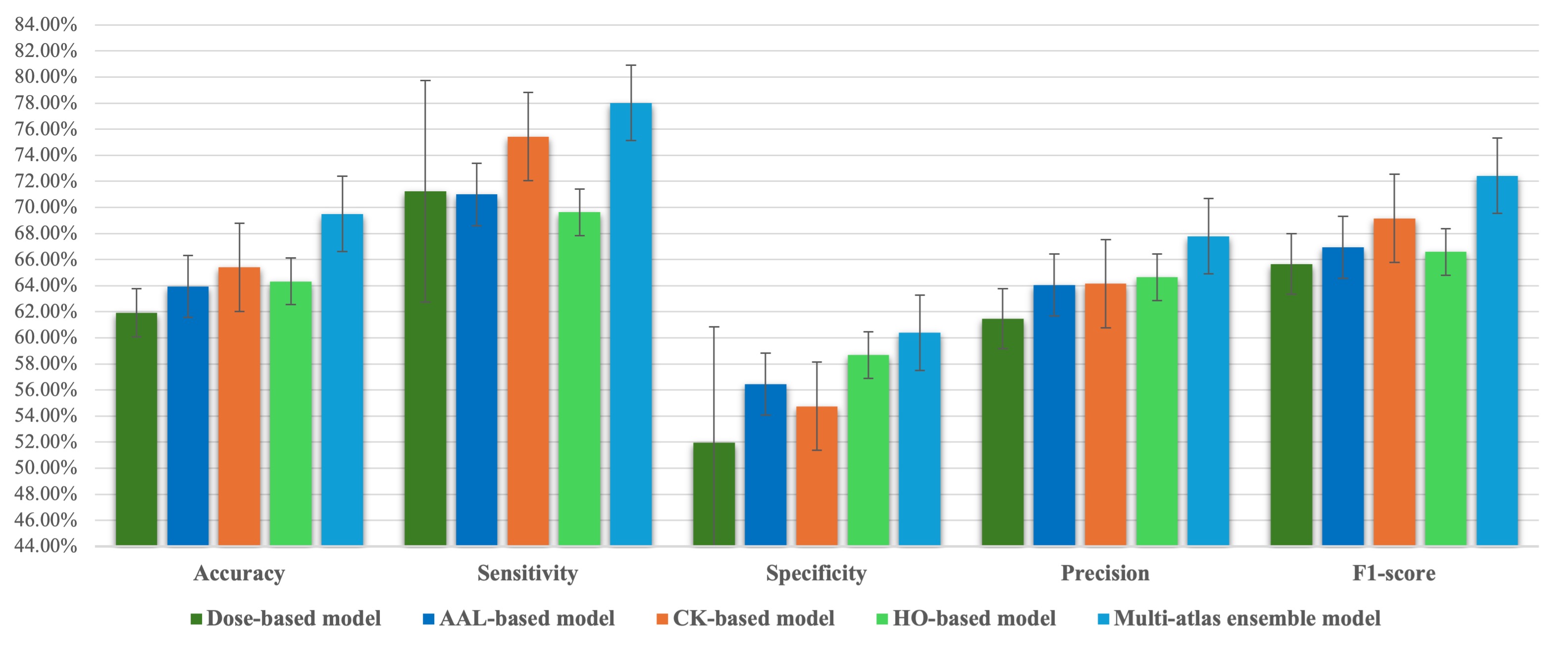}
    \caption{Comparison of Classification Performance Metrics Between the Majority Voting Multi-Atlas Ensemble GAT Model and Single-Atlas GAT Models.}
    \label{fig:Subfigure 2}
\end{figure}

Accordingly, we chose a majority voting ensemble since it outperformed both sum-based and weighted sum-based models. Moreover, we compared the performance of the multi-atlas model based on a majority voting with that of other single-atlas models in order to obtain further clarity on its prediction performance. Based on Tables \ref{SingleAt_SMOTE}, \ref{ResultsEnsemble} and Figure \ref{fig:Subfigure 2}, the multi-atlas model shows better prediction performance than the four models that relied on a single atlas. The multi-atlas model improves all metrics when compared with the Dose-based model: accuracy by 7.58\%, sensitivity by 6.79\%, specificity by 8.42\%, precision by 6.32\%, and F1-score by 6.77\%. The multi-atlas model also improves accuracy by 5.54\%, sensitivity by 7.03\%, specificity by 3.94\%, precision by 3.75\%, and F1-score by 5.48\% over the AAL-based model. A comparison with the CK-based model reveal an improvement in accuracy, sensitivity, specificity, precision, and F1-score by 4.08\%, 2.59\%, 5.65\%, 3.64\%, and 3.27\%, respectively. Additionally, the accuracy, sensitivity, specificity, precision, and F1-score of the multi-atlas model are enhanced over the HO-based model by 5.16\%, 8.39\%, 1.71\%, 3.14\%, and 5.84\%, respectively.
\begin{table}[h]
\caption{Average Results and Standard Deviation from 10-Fold Stratified Cross-Validation of the Multi-Atlas Ensemble GAT Model with SMOTE Across Different Ensembling Approaches.\\}
\label{ResultsEnsemble}
\resizebox{\columnwidth}{!}{%
\renewcommand{\arraystretch}{1.5}
\begin{tabular}{|c|c|c|c|c|c|}
\hline \multirow{2}{*}{\textbf{Method}} & \multicolumn{5}{c|}{ \textbf{Average $\pm$ Standard deviation (\%)} } \\
\cline { 2 - 6 } & \textbf{Accuracy} & \textbf{Sensitivity} & \textbf{Specificity} & \textbf{Precision} & \textbf{F1-score} \\
\hline 
\begin{tabular}{c}Majority voting
\end{tabular} & $\mathbf{69.49\pm3.54}$ & $\mathbf{78.02\pm6.55}$ & $\mathbf{60.39\pm5.41}$ & $\mathbf{67.79\pm2.87}$ & $\mathbf{72.43\pm3.71}$ \\
\hline \begin{tabular}{c}Sum\end{tabular}& $67.83\pm3.66$ & $77.78\pm6.51$ & $57.24\pm8.71$ & $66.24\pm3.77$ & $71.34\pm3.28$   \\
\hline \begin{tabular}{c} 
Weighted sum
\end{tabular} & $67.45\pm3.20$ & $77.65\pm6.33$ & $56.58\pm8.40$ & $65.84\pm3.45$ & $71.06\pm2.94$ \\
\hline
\end{tabular}%
}
{\\\\The bolded values indicate the best performing results.}
\end{table}
\begin{table}[h]
\caption{Classification Results for the Fold with the Highest Accuracy from 10-Fold Stratified Cross-Validation of the Multi-Atlas Ensemble GAT Model with SMOTE Across Different Ensembling Approaches.\\}
\label{BestEnsemble}
\resizebox{\columnwidth}{!}{%
\renewcommand{\arraystretch}{1.5}
\begin{tabular}{|c|c|c|c|c|c|}
\hline \textbf{Method} & \textbf{Accuracy (\%)} & \textbf{Sensitivity (\%)} & \textbf{Specificity (\%)}& \textbf{Precision (\%)} & \textbf{F1-score (\%)} \\
\hline \begin{tabular}{c}
Majority voting
\end{tabular} & $\mathbf{75.80}$ & $\mathbf{88.89}$ & $\mathbf{61.84}$ & $\mathbf{71.29}$ & $\mathbf{79.12}$  \\
\hline \begin{tabular} {c}
Sum 
\end{tabular} & $73.25$ & $85.19$ & $60.53$ & $69.70$ & $76.67$ \\
\hline \begin{tabular} {c}
Weighted sum
\end{tabular} & $70.70$ & $82.72$ & $57.89$ & $67.68$ & $74.44$ \\
\hline
\end{tabular}%
}
{\\ \\ The bolded values indicate the best performing results.}
\end{table}

\subsubsection{Statistical Analysis}
In this section, we examine if the multi-atlas model differs significantly from each single-atlas model using an independent two-sample t-test, as illustrated in Table~\ref{ReslutsTTest}. In this study, we assumed that differences between models were statistically significant when $p < 0.05$. The results show that there is a significant difference between the multi-atlas model and the Dose-based model on four metrics, including accuracy, specificity, precision, and F1-score. Meanwhile, the accuracy, sensitivity, precision, and F1-score of our multi-atlas model differ significantly from those of the AAL-based model and the HO-based model. Furthermore, the CK-based model and the multi-atlas model show significant differences on both accuracy and precision.
\begin{table}[h]
\caption{Comparison of p-values Between the Majority Voting Multi-Atlas Ensemble GAT Model and Single-Atlas GAT Models Using SMOTE.\\}
\label{ReslutsTTest}
\resizebox{\columnwidth}{!}{%
\renewcommand{\arraystretch}{1.6}
\begin{tabular}{|c|c|c|c|c|c|}
\hline \textbf{Compared models} & \textbf{Accuracy} & \textbf{Sensitivity} & \textbf{Specificity} & \textbf{Precision} & \textbf{F1-score} \\
\hline \begin{tabular}{c} 
Multi-atlas model and Dose-based model
\end{tabular} & $\mathbf{0.0000}$ & 0.0738 & $\mathbf{0.0257}$ & $\mathbf{0.0001}$ & $\mathbf{0.0007}$ \\
\hline \begin{tabular}{c} 
Multi-atlas model and AAL-based model
\end{tabular} & $\mathbf{0.0061}$ & $\mathbf{0.0447}$ & 0.3762 & $\mathbf{0.0439}$ & $\mathbf{0.030}$ \\
\hline \begin{tabular}{c} 
Multi-atlas model and CK-based model
\end{tabular} & $\mathbf{0.0158}$ & 0.4019 & 0.0748 & $\mathbf{0.0156}$ & 0.0519 \\
\hline \begin{tabular}{c} 
Multi-atlas model and HO-based model
\end{tabular} & $\mathbf{0.0006}$ & $\mathbf{0.0376}$ & 0.6594 & $\mathbf{0.0282}$ & $\mathbf{0.0014}$ \\
\hline
\end{tabular}%
}
{\\ \\ The bolded values indicate statistically significant differences between models at $p < 0.05$.}
\end{table}
\subsubsection{Performance Comparison of GAT Models With and Without SMOTE}
SMOTE is a powerful technique for dealing with class imbalance, which has achieved robust results in a variety of applications. In SMOTE, synthetic samples are added to the minority class to create a balanced dataset \citep{chawla2002smote}. This study utilized SMOTE to oversample MDD and HC classes in both the training and validation sets, so that only real data was included in the testing set. Table \ref{CompareWithoutSMOTE} presents an analysis of SMOTE's effectiveness on MDD prediction using the proposed multi-atlas model and single-atlas models trained on a multi-site dataset. Based on the results, the SMOTE technique has effectively improved the performance of classification models. Specifically, the Dose-based model shows improvements of 0.06\%, 2.09\%, and 1.0\% in accuracy, sensitivity, and F1-score. In the AAL-based model, sensitivity and F1-score are improved by 4.45\% and 1.39\%, respectively. Meanwhile, the sensitivity and F1-score of the CK-based model are increased by 2.71\% and 0.78\%, respectively. In contrast, the HO-based model is only improved in terms of specificity by 3.42\%. Moreover, the multi-atlas model achieves improvement in accuracy, sensitivity, and F1-score by 0.57\%, 3.7\%, and 1.3\%, respectively. Clearly, the proposed multi-atlas model based on SMOTE has superior performance compared to the other models. This model gives an accuracy of 69.49±3.54\%, a sensitivity of 78.02±6.55\%, a specificity of 60.39±5.41\%, a precision of 67.79±2.87\%, and an F1-score of 72.43±3.71\%. However, the statistical significance analysis indicates no statistically significant differences between using SMOTE and without it with $p < 0.05$.
\begin{table}[h]
 \caption{Performance Comparison Between the Multi-Atlas Ensemble GAT Model and Single-Atlas GAT Models with and without SMOTE Using Multi-Site Data.\\}
\label{CompareWithoutSMOTE}
\renewcommand{\arraystretch}{1.7}
\begin{tabular}{|cc|c|c|c|c|c|}
\hline 
\multicolumn{2}{|c}{\multirow{2}{*}{\textbf{Method}}} & \multicolumn{5}{|p{350pt}|}{ \centering \textbf{Average} $\pm$ \textbf{Standard deviation (\%)} } \\
\cline{3-7} & & \centering \textbf{Accuracy} & \centering \textbf{Sensitivity} & \centering \textbf{Specificity} & \centering \textbf{Precision} & \centering\arraybackslash \textbf{F1-score}\\
\hline 
\multicolumn{1}{|c}{\multirow{8}{*}{ \rotatebox[origin=c]{90}{Without SMOTE}}}& \multicolumn{1}{|p{60pt}}{\centering Dose-based model} & \multicolumn{1}{|c}{ \begin{tabular}{c} 61.85$\pm$ 1.72\end{tabular}} & \multicolumn{1}{|c}{\begin{tabular}{c} 69.14$\pm$ 12.28\end{tabular}} & \multicolumn{1}{|c}{\begin{tabular}{c} 54.08$\pm$ 15.26\end{tabular}} & \multicolumn{1}{|c}{\begin{tabular}{c} 62.72$\pm$ 4.84\end{tabular}} & \multicolumn{1}{|c|}{\begin{tabular}{c} 64.67$\pm$ 4.12\end{tabular}}\\ 
\cline{2-7} 
 & \multicolumn{1}{|p{60pt}}{\centering AAL-based model} & \multicolumn{1}{|c}{\begin{tabular}{c} 64.46$\pm$2.38\end{tabular}} & \multicolumn{1}{|c}{\begin{tabular}{c} 66.54$\pm$9.59\end{tabular}} & \multicolumn{1}{|c}{\begin{tabular}{c}62.24$\pm$9.23\end{tabular}} & \multicolumn{1}{|c}{\begin{tabular}{c}65.69$\pm$3.25\end{tabular}} & \multicolumn{1}{|c|}{\begin{tabular}{c} 65.56$\pm$4.53\end{tabular}}\\
\cline{2-7} 
& \multicolumn{1}{|p{60pt}}{\centering CK-based model} & \multicolumn{1}{|c}{\begin{tabular}{c}65.54$\pm$2.71\end{tabular}} & \multicolumn{1}{|c}{\begin{tabular}{c} 72.72$\pm$8.51\end{tabular}} & \multicolumn{1}{|c}{\begin{tabular}{c} 57.89$\pm$11.33\end{tabular}} & \multicolumn{1}{|c}{\begin{tabular}{c} 65.37$\pm$4.15\end{tabular}} & \multicolumn{1}{|c|}{\begin{tabular}{c}68.38$\pm$2.79\end{tabular}}\\ 
\cline{2-7} 
& \multicolumn{1}{|p{60pt}}{\centering HO-based model} & \multicolumn{1}{|c}{\begin{tabular}{c} 66.43$\pm$1.14\end{tabular}} & \multicolumn{1}{|c}{\begin{tabular}{c} 76.91$\pm$7.20\end{tabular}} & \multicolumn{1}{|c}{\begin{tabular}{c}55.26$\pm$8.63\end{tabular}} & \multicolumn{1}{|c}{\begin{tabular}{c}65.00$\pm$2.58\end{tabular}} & \multicolumn{1}{|c|}{\begin{tabular}{c} 70.16$\pm$1.93\end{tabular}}\\ 
\cline{2-7} 
& \multicolumn{1}{|p{60pt}}{\centering Multi-atlas model} & \multicolumn{1}{|c}{\begin{tabular}{c} 68.92$\pm$1.86\end{tabular}} & \multicolumn{1}{|c}{\begin{tabular}{c}74.32$\pm$3.86\end{tabular}} & \multicolumn{1}{|c}{\begin{tabular}{c} {$\mathbf{63.16\pm}$$\mathbf{4.52}$}\end{tabular}} & \multicolumn{1}{|c}{\begin{tabular}{c} {$\mathbf{68.35\pm}$$\mathbf{2.16}$}\end{tabular}} & \multicolumn{1}{|c|}{\begin{tabular}{c}71.13$\pm$1.90\end{tabular}}\\ 
\hline 
\multicolumn{1}{|c}{\multirow{8}{*}{\rotatebox[origin=c]{90} {With SMOTE}}} & \multicolumn{1}{|p{60pt}}{\centering Dose-based model} & \multicolumn{1}{|c}{\begin{tabular}{c}61.91$\pm$1.84\end{tabular}} & \multicolumn{1}{|c}{\begin{tabular}{c}71.23$\pm$8.50\end{tabular}} & \multicolumn{1}{|c}{\begin{tabular}{c}51.97$\pm$8.87\end{tabular}} & \multicolumn{1}{|c}{\begin{tabular}{c}61.47$\pm$2.29\end{tabular}} & \multicolumn{1}{|c|}{\begin{tabular}{c}65.66$\pm$3.32\end{tabular}}\\ 
\cline{2-7} 
& \multicolumn{1}{|p{60pt}}{\centering AAL-based model} & \multicolumn{1}{|c}{\begin{tabular}{c} 63.95$\pm$4.01\end{tabular}} & \multicolumn{1}{|c}{\begin{tabular}{c} 70.99$\pm$7.27\end{tabular}} & \multicolumn{1}{|c}{\begin{tabular}{c} 56.45$\pm$11.88\end{tabular}} & \multicolumn{1}{|c}{\begin{tabular}{c} 64.04$\pm$4.33\end{tabular}} & \multicolumn{1}{|c|}{\begin{tabular}{c} 66.95$\pm$3.03\end{tabular}}\\ 
\cline{2-7}
 & \multicolumn{1}{|p{60pt}}{\centering CK-based model} & \multicolumn{1}{|c}{\begin{tabular}{c} 65.41$\pm$2.92\end{tabular}} & \multicolumn{1}{|c}{\begin{tabular}{c}75.43$\pm$6.26\end{tabular}} & \multicolumn{1}{|c}{\begin{tabular}{c}54.74$\pm$7.16\end{tabular}} & \multicolumn{1}{|c}{\begin{tabular}{c}64.15$\pm$2.91\end{tabular}} & \multicolumn{1}{|c|}{\begin{tabular}{c} 69.16$\pm$2.90\end{tabular}}\\ 
\cline{2-7} 
& \multicolumn{1}{|p{60pt}}{\centering HO-based model} & \multicolumn{1}{|c}{\begin{tabular}{c}64.33$\pm$1.07\end{tabular}} & \multicolumn{1}{|c}{\begin{tabular}{c}69.63$\pm$9.11\end{tabular}} & \multicolumn{1}{|c}{\begin{tabular}{c}58,68$\pm$10.09\end{tabular}} & \multicolumn{1}{|c}{\begin{tabular}{c} 64.65$\pm$2.70\end{tabular}} & \multicolumn{1}{|c|}{\begin{tabular}{c}66.59$\pm$2.78\end{tabular}}\\ 
\cline{2-7} 
 & \multicolumn{1}{|p{60pt}}{\centering Multi-atlas model} & \multicolumn{1}{|c}{\begin{tabular}{c}{$\mathbf{69.49\pm}$$\mathbf{3.54}$}\end{tabular}} & \multicolumn{1}{|c}{\begin{tabular}{c}{$\mathbf{78.02\pm}$$\mathbf{6.55}$}\end{tabular}} & \multicolumn{1}{|c}{\begin{tabular}{c}60.39$\pm$5.41\end{tabular}} & \multicolumn{1}{|c}{\begin{tabular}{c}$67.79\pm2.87$\end{tabular}} & \multicolumn{1}{|c|}{\begin{tabular}{c} {$\mathbf{72.43\pm}$$\mathbf{3.71}$}\end{tabular}}\\ 
\hline 
\end{tabular}
{\\\\The bolded values indicate the best performing results.}
\end{table}

\subsubsection{Performance Comparison With Other Models Using Single Site Data}
In this section, we evaluated our proposed approach on data collected from a single site (site 20), which has the largest number of subjects among the REST-meta-MDD sites, as shown in Table \ref{CompareWithoutSingleSite}. It consisted of 470 subjects, of which 245 had MDD and 225 had HC \citep{chen2022direct}\citep{yan2019reduced}. Based on the results, the multi-atlas model without SMOTE has superiority in four metrics: an accuracy of 71.06$\pm$5.40\%, a sensitivity of 86.00$\pm$5.73\%, a precision of 68.37$\pm$4.98\%, and an F1-score of 76.01$\pm$4.05\%. In contract, the multi-atlas model based on SMOTE achieves the second highest performance, with 69.79$\pm$4.74\% accuracy, 84.80$\pm$6.14\% sensitivity, 52.73$\pm$11.89\% specificity, 67.59$\pm$5.24\% precision, and 74.95$\pm$3.26\% F1-score. Compared to single-atlas models, the CK-based model with SMOTE achieves the best performance, with an accuracy of 68.09$\pm$3.81\%, a sensitivity of 84.00$\pm$7.16\%, a specificity of 50.00$\pm$11.13\%, a precision of 66.02$\pm$4.22\%, and an F1-score of 73.64$\pm$3.00\%. Similarly, the Dose-based model shows good performance with and without SMOTE. It delivers 67.45$\pm$4.37\% accuracy, 82.40$\pm$8.04\% sensitivity, 50.45$\pm$16.3\% specificity, 66.33$\pm$5.58\% precision, and 72.93$\pm$2.22\% F1-score. The AAL-based model is improved with SMOTE to produce 61.70$\pm$5.21\% accuracy, 70.40$\pm$8.98\% Sensitivity, 51.82$\pm$17.74\% specificity, 63.59$\pm$6.47\% precision, and 66.12$\pm$3.18\% F1-score. However, the HO-based model with SMOTE achieves the lowest performance with 57.23$\pm$2.2\% accuracy, 66.40$\pm$12.67\% sensitivity, 46.82$\pm$13.9\% specificity, 59.05$\pm$2.44\% precision, and 61.72$\pm$5.37\% F1-score.

\begin{table}[h]
\caption{Performance Comparison Between the Multi-Atlas Ensemble GAT Model and Single-Atlas GAT Models with and without SMOTE Using Single-Site Data.\\}
\label{CompareWithoutSingleSite}
\resizebox{\textwidth}{!}{%
\renewcommand{\arraystretch}{1.7}
\begin{tabular}{|cc|c|c|c|c|c|}
\hline 
\multicolumn{2}{|c}{\multirow{2}{*}{\textbf{Method}}} & \multicolumn{5}{|c|}{ \centering \textbf{Average} $\pm$ \textbf{Standard deviation (\%)} } \\
\cline{3-7} & & \centering \textbf{Accuracy} & \centering \textbf{Sensitivity} & \centering \textbf{Specificity} & \centering \textbf{Precision} & \centering\arraybackslash \textbf{F1-score} \\ 
\hline 
\multicolumn{1}{|c}{\multirow{8}{*}{\rotatebox[origin=c]{90}{Without SMOTE}}} & \multicolumn{1}{|p{60pt}}{\centering Dose-based model} & \multicolumn{1}{|c}{ \begin{tabular}{c}67.45$\pm$4.37 \end{tabular}} & \multicolumn{1}{|c}{\begin{tabular}{c}82.40$\pm$8.04 \end{tabular}} & \multicolumn{1}{|c}{\begin{tabular}{c} 50.45$\pm$16.32 \end{tabular}} & \multicolumn{1}{|c}{\begin{tabular}{c}66.33$\pm$5.58 \end{tabular}} & \multicolumn{1}{|c|}{\begin{tabular}{c}72.93$\pm$2.22 \end{tabular}}\\ 
\cline{2-7} 
& \multicolumn{1}{|p{60pt}}{\centering AAL-based model} & 
\multicolumn{1}{|c}{\begin{tabular}{c}58.94$\pm$3.69\end{tabular}} & 
\multicolumn{1}{|c}{\begin{tabular}{c}79.60$\pm$17.39 \end{tabular}} &
\multicolumn{1}{|c}{\begin{tabular}{c}35.45$\pm$23.95 \end{tabular}} & 
\multicolumn{1}{|c}{\begin{tabular}{c} 59.77$\pm$5.39 \end{tabular}} &
\multicolumn{1}{|c|}{\begin{tabular}{c} 66.59$\pm$5.74 \end{tabular}}\\
\cline{2-7} 
& \multicolumn{1}{|p{60pt}}{\centering CK-based model} & \multicolumn{1}{|c}{\begin{tabular}{c}67.23$\pm$5.05 \end{tabular}} & 
\multicolumn{1}{|c}{\begin{tabular}{c} 80.00$\pm$6.45 \end{tabular}} & 
\multicolumn{1}{|c}{\begin{tabular}{c} 52.73$\pm$9.79 \end{tabular}} & 
\multicolumn{1}{|c}{\begin{tabular}{c} 66.06$\pm$4.52 \end{tabular}} & 
\multicolumn{1}{|c|}{\begin{tabular}{c} 72.19$\pm$3.97 \end{tabular}}\\ 
\cline{2-7} 
& \multicolumn{1}{|p{60pt}}{\centering HO-based model} & 
\multicolumn{1}{|c}{\begin{tabular}{c} 62.55$\pm$4.06 \end{tabular}} & 
\multicolumn{1}{|c}{\begin{tabular}{c} 64.80$\pm$15.68 \end{tabular}} & \multicolumn{1}{|c}{\begin{tabular}{c} {$\mathbf{60.00\pm}$$\mathbf{19.58}$} \end{tabular}} & 
\multicolumn{1}{|c}{\begin{tabular}{c} 67.11$\pm$7.87 \end{tabular}} & 
\multicolumn{1}{|c|}{\begin{tabular}{c} 63.81$\pm$8.03 \end{tabular}}\\ 
\cline{2-7} 
& \multicolumn{1}{|p{60pt}}{\centering Multi-atlas model} & \multicolumn{1}{|c}{\begin{tabular}{c}{$\mathbf{71.06\pm}$$\mathbf{5.40}$} \end{tabular}} & 
\multicolumn{1}{|c}{\begin{tabular}{c} {$\mathbf{86.00 \pm}$$\mathbf{5.73}$} \end{tabular}} & 
\multicolumn{1}{|c}{\begin{tabular}{c} 54.00$\pm$10.84\end{tabular}} & 
\multicolumn{1}{|c}{\begin{tabular}{c}{$\mathbf{68.37\pm}$$\mathbf{4.98}$} \end{tabular}} & 
\multicolumn{1}{|c|}{\begin{tabular}{c} {$\mathbf{76.01\pm}$$\mathbf{4.05}$} \end{tabular}}\\
\hline 
\multicolumn{1}{|c}{\multirow{8}{*}{\rotatebox[origin=c]{90} {With SMOTE}}} & \multicolumn{1}{|p{60pt}}{\centering Dose-based model} & \multicolumn{1}{|c}{\begin{tabular}{c} 67.45$\pm$3.81\end{tabular}} & \multicolumn{1}{|c}{\begin{tabular}{c} 76.00$\pm$10.28\end{tabular}} & \multicolumn{1}{|c}{\begin{tabular}{c}57.73$\pm$11.33\end{tabular}} & \multicolumn{1}{|c}{\begin{tabular}{c} 67.60$\pm$4.09\end{tabular}} & \multicolumn{1}{|c|}{\begin{tabular}{c} 71.02$\pm$4.74\end{tabular}}\\ 
\cline{2-7}
& \multicolumn{1}{|p{60pt}}{\centering AAL-based model} & \multicolumn{1}{|c}{\begin{tabular}{c} 61.70$\pm$5.21\end{tabular}} & \multicolumn{1}{|c}{\begin{tabular}{c}70.40$\pm$8.98\end{tabular}} & \multicolumn{1}{|c}{\begin{tabular}{c}51.82$\pm$17.74\end{tabular}} & \multicolumn{1}{|c}{\begin{tabular}{c} 63.59$\pm$6.47\end{tabular}} & \multicolumn{1}{|c|}{\begin{tabular}{c}66.12$\pm$3.18\end{tabular}}\\ 
\cline{2-7}
& \multicolumn{1}{|p{60pt}}{\centering CK-based model} & \multicolumn{1}{|c}{\begin{tabular}{c}68.09$\pm$3.81\end{tabular}} & 
\multicolumn{1}{|c}{\begin{tabular}{c}84.00$\pm$7.16\end{tabular}} & 
\multicolumn{1}{|c}{\begin{tabular}{c}50.00$\pm$11.13\end{tabular}} & 
\multicolumn{1}{|c}{\begin{tabular}{c} 66.02$\pm$4.22\end{tabular}} & 
\multicolumn{1}{|c|}{\begin{tabular}{c}73.64$\pm$3.00\end{tabular}}\\ 
\cline{2-7} 
& \multicolumn{1}{|p{60pt}}{\centering HO-based model} & \multicolumn{1}{|c}{\begin{tabular}{c}57.23$\pm$2.22\end{tabular}} & 
\multicolumn{1}{|c}{\begin{tabular}{c}66.40$\pm$12.67\end{tabular}} & \multicolumn{1}{|c}{\begin{tabular}{c} 46.82$\pm$13.94\end{tabular}} & 
\multicolumn{1}{|c}{\begin{tabular}{c} 59.05$\pm$2.44 \end{tabular}} & 
\multicolumn{1}{|c|}{\begin{tabular}{c} 61.72$\pm$5.37\end{tabular}}\\ 
\cline{2-7} 
& \multicolumn{1}{|p{60pt}}{\centering Multi-atlas model} & \multicolumn{1}{|c}{\begin{tabular}{c}69.79$\pm$4.74\end{tabular}} & 
\multicolumn{1}{|c}{\begin{tabular}{c}84.80$\pm$6.14\end{tabular}} & \multicolumn{1}{|c}{\begin{tabular}{c}52.73$\pm$11.89\end{tabular}} & 
\multicolumn{1}{|c}{\begin{tabular}{c} 67.59$\pm$5.24\end{tabular}} &
\multicolumn{1}{|c|}{\begin{tabular}{c}74.95$\pm$3.26\end{tabular}}\\ 
\hline 
\end{tabular}%
}
{\\\\The bolded values indicate the best performing results.}
\end{table}

As a matter of fact, it could be argued that the differences in SMOTE's effectiveness between single-site and multi-site data can be attributed to their distinctive characteristics. When using SMOTE on single-site data, where imaging protocols and populations are homogeneous, synthetic samples brought into the model may not add significant diversity, resulting in an overall reduced performance of 0.78\% to 1.27\% in all metrics. Conversely, multi-site data have a higher level of variability due to differences in the acquisition protocols, scanner types, and demographic characteristics of the population, resulting in more complex and integrated distributions. In this context, SMOTE is beneficial since it generates synthetic samples that reduce overfitting and balance data across sites. However, there is no doubt that the multi-atlas models have a better performance than the single-atlas models. Moreover, the evaluation metrics for the fold that has the highest accuracy for each multi-atlas model are displayed in Table \ref{BestSinglesite}. Without SMOTE, the best multi-atlas model among all 10 folds based on single-site data provides an accuracy of 80.85\%, a sensitivity of 92.00\%, a specificity of 68.18\%, a precision of 76.67\%, and an F1-score of 83.64\%. In contrast, the best performing multi-atlas model based on multi-site data has 71.97\% accuracy, 79.01\% sensitivity, 64.47\% specificity, 70.33\% precision, and 74.42\% F1-score. Using SMOTE, the best multi-atlas model based on single-site data shows accuracy of 76.60\%, sensitivity of 76.00\%, specificity of 77.27\%, precision of 79.17\%, and F1 score of 77.55\%. While the best performing model based on multi-site data achieves 75.80\% accuracy, 88.89\% sensitivity, 61.84\% specificity, 71.26\% precision, and 79.12\% F1-score.
\begin{table}[h]
\caption{Classification Results for the Fold with the Highest Accuracy from 10-Fold Stratified Cross-Validation of the Multi-Atlas Ensemble GAT Models.\\}
\label{BestSinglesite}
\resizebox{\columnwidth}{!}{%
\renewcommand{\arraystretch}{2}
\begin{tabular}{|c|c|c|c|c|c|c|}
\hline \centering \textbf{Method} & \centering \textbf{Site} & \textbf{Accuracy (\%)} & \textbf{Sensitivity (\%)}& \textbf{Specificity (\%)} & \textbf{Precision (\%)} & \textbf{F1-score (\%)} \\
\hline 
\multicolumn{1}{|c|}{\multirow{2}{*}{{Without SMOTE}}}& Single-site &$\mathbf{80.85}$ & $\mathbf{92.00}$ & $68.18$ & $76.67$ & $\mathbf{83.64}$ \\
\cline{2-7} 
& Multi-site & 71.97 & 79.01 & 64.47 & 70.33 & 74.42  \\
\hline 
\multicolumn{1}{|c|}{\multirow{2}{*}{{With SMOTE}}}& Single-site & $76.60$ & $76.00$ & $\mathbf{77.27}$ & $\mathbf{79.17}$ & $77.55$ \\
\cline{2-7} 
&  Multi-site& 75.80  & 88.89  & 61.84  & 71.26 & 79.12  \\
\hline
\end{tabular}%
}
{\\\\The bolded values indicate the best performing results.}
\end{table}

\subsubsection{Performance Comparison With The Existing Studies}
A performance comparison of the proposed model with other existing models is provided in Table \ref{ReslutsWithOtherModels}. We conducted a comparative experiment with models developed by \citep{liu2023spatial}, \citep{xia2023depressiongraph}, and \citep{lee2024spectral}, which utilized the same dataset as this study. We compared these models with our multi-site and single-site models with and without SMOTE. Models by \citep{xia2023depressiongraph} and \citep{lee2024spectral} differ from ours in that they have been validated five times, use single-site data, and do not use oversampling. Comparatively, our models and those published in \citep{liu2023spatial} used a 10-fold validation approach, data from multiple sites, and oversampling techniques. Further, we have highlighted the studies that utilized an oversampling technique with an asterisk (*) next to them in the Table \ref{ReslutsWithOtherModels}. The results indicate that the model proposed by \citep{xia2023depressiongraph} achieves the best performance in three metrics: an accuracy of 71.48\%, a sensitivity of 95.04\%, and an F1-score of 77.91\%. Actually, this model was evaluated based on data obtained from only a single site (site 20). In contrast, our multi-atlas model without SMOTE and trained on single-site data (site 20) achieved the second best performance, with 71.06$\pm$5.40\% accuracy, 86.00$\pm$5.73\% sensitivity, 54.00$\pm$10.84\% specificity, and 76.01$\pm$4.05\%. It also exhibits a precision of 68.37$\pm$4.98\%, which is higher than that of other single-site models. Following this, our multi-atlas model with SMOTE produces an accuracy of 69.79$\pm$4.74\%, a sensitivity of 84.80$\pm$6.14\%, a specificity of 52.73$\pm$11.89\%, a precision of 67.59$\pm$5.24\%, and an F1-score of 74,95$\pm$3.26\%. The model proposed by \citep{lee2024spectral} was evaluated using data from a single site (site 20), which yielded an accuracy of 68.88±2.55\%, a sensitivity of 69.98±5.97\%, and an F1-score of 70.02±2.79\%. However, this model exhibits a specificity of 67.62±6.81\%, which is higher than that of all competing models. 

Moreover, the proposed multi-atlas model with SMOTE was evaluated using data from 16 sites and demonstrated superior performance over other multi-site models. It achieved 69.49±3.54\% accuracy, 78.02±6.55\% sensitivity, 60.39±5.41\% specificity, 68.35\% precision, and 72.43±3.71\% F1-score. Particularly, the model showed improvements in accuracy between 0.57\% and 6.08\%, sensitivity between 3.7\% and 17.45\%, and F1-score between 1.3\% and 11.1\% compared to other multi-site models. Meanwhile, our proposed multi-atlas model without SMOTE achieved the second best performance among multi-sites models, with 68.92$\pm$1.86\% accuracy, 74.32$\pm$3.86\% sensitivity, and 71.13$\pm$1.90\% F1-score. The specificity and precision of this model are 63.16$\pm$4.52\% and 68.35$\pm$2.16\%, which are higher than those of the multi-site models. In \citep{liu2023spatial}, a spatial-temporal data-augmentation-based classification model was developed using data from 24 sites. Compared to other models, it attains the lowest performance with an accuracy of 63.41\%, a sensitivity of 60.57\% for MDD and 66.13\% for HC, a precision of 62.53\% for MDD and 64.17\% for HC, and an F1-score of 61.33\%.   

\begin{table}[h]
\caption{Performance Comparison Between the Proposed Model and Other Existing Models.\\}
\label{ReslutsWithOtherModels}
\resizebox{\textwidth}{!}{
\renewcommand{\arraystretch}{1.8}
\begin{tabular}{|c|c|c|c|c|c|c|c|c|}
\hline
\multicolumn{1}{|c|}{\multirow{2}{*}{ \textbf{Site}}} 
& \multirow{2}{*}{\textbf{Method}} 
& \multirow{2}{*}{\begin{tabular}{c}\textbf{Number of} \\ \textbf{subjects} \end{tabular} } & \multirow{2}{*}{\textbf{Atlas}} & \multicolumn{5}{c|}{ \textbf{Average $\pm$ Standard deviation (\%)} } \\
\cline{5-9} & & & &\centering \textbf{ Accuracy} & \centering \textbf{Sensitivity} & \centering \textbf{Specificity} & \centering \textbf{Precision} & \centering\arraybackslash \textbf{ F1-score} \\
\hline
\multicolumn{1}{|c|}{\multirow{7}{*}{\rotatebox[origin=c]{90} { Single-site}}} & 
\centering \citep{xia2023depressiongraph} & 
\begin{tabular}{c} 533\end{tabular} & 
\begin{tabular}{c} AAL, CK, \\HO \end{tabular} & $\mathbf{71.48}$ & $\mathbf{95.04}$ & $45.02$ & $66.01$ & $\mathbf{77.91}$ \\
\cline{2-9} 
&
\centering \citep{lee2024spectral} & \begin{tabular}{c}
{470}\end{tabular} & 
\begin{tabular}{c} AAL, CK, \\HO \end{tabular} & 
${68.88\pm}$${2.55}$ & ${69.98\pm}$${5.97}$ & \begin{tabular}{c}
$\mathbf{67.62\pm}$$\mathbf{6.81}$ \end{tabular} & \begin{tabular}{c} ------ \end{tabular} & ${70.02\pm}$${2.79}$ \\
\cline{2-9}  & 
\begin{tabular}{c}Without SMOTE \\ (Ours) \end{tabular}  & 
\begin{tabular}{c} 470 \end{tabular} &
\begin{tabular}{c} Dose, AAL, \\CK, HO
\end{tabular}& 
${71.06\pm}$${5.40}$ & ${86.00\pm}$${5.73}$ & ${54.09\pm}$${10.84}$ & $\mathbf{68.37\pm}$$\mathbf{4.98}$ & ${76.01\pm}$${4.05}$ \\
\cline{2-9} &
\begin{tabular}{c}With SMOTE \\ (Ours)* \end{tabular} &
\begin{tabular}{c} 940\end{tabular}
&\begin{tabular}{c} Dose, AAL, \\ CK, HO \end{tabular} 
& ${69.79\pm}$${4.74}$ & ${84.80\pm}$${6.14}$ & ${52.73\pm}$${11.89}$ & 
\begin{tabular}{c}${67.59\pm}$${5.24}$ \end{tabular}& ${74.95\pm}$${3.26}$ \\
\hline
\multicolumn{1}{|c|}{\multirow{5}{*}{ \rotatebox[origin=c]{90} { Multi-site}}} & 
\centering \citep{liu2023spatial}* & 
\begin{tabular}{c} 2379 \end{tabular} & 
\begin{tabular}{c}AAL \end{tabular} & $63.41$ & 
\begin{tabular}{c} 
$\mathrm{MDD=}\ 60.57$ \\
$\mathrm{HC=}\ 66.13$ \end{tabular} &
\begin{tabular}{c} ------ \end{tabular} &
\begin{tabular}{c}
$\mathrm{MDD=}\ 62.53$ \\
$\mathrm{HC=}\ 64.17$ \end{tabular} & $61.33$ \\
\cline{2-9} &
\begin{tabular}{c}Without SMOTE \\ (Ours) \end{tabular} 
& 
\begin{tabular}{c} {1563} \end{tabular} & 
\begin{tabular}{c} Dose, AAL, \\ CK, HO \end{tabular} & ${68.92\pm}$${1.86}$ & ${74.32\pm}$${3.86}$ & $\mathbf{63.16\pm4.52}$ & $\mathbf{68.35\pm2.16}$ & ${71.13\pm}$${1.90}$ \\
\cline{2-9}  &
\begin{tabular}{c}With SMOTE \\ (Ours)* \end{tabular} 
& 
\begin{tabular}{c} 2969 \end{tabular}
&\begin{tabular}{c} Dose, AAL, \\CK, HO \end{tabular} & 
$\mathbf{69.49\pm3.54}$ & 
$\mathbf{78.02\pm6.55}$ & ${60.39\pm}$${5.41}$ & 
\begin{tabular}{c}$67.79\pm2.87$ \end{tabular} & $\mathbf{72.43\pm3.71}$ \\
\hline
\end{tabular}
}
{\\\\ \textasteriskcentered \ Methods were conducted using the oversampling technique.\\The best performing results are bolded for each site category.}
\end{table}

\section{Conclusion}
In this study, we developed an ensemble-based GNN model for the classification of MDD based on rs-fMRI data. Our ensemble model was constructed by combining features derived from four brain segmentation atlases to capture brain complexity and identify distinct features more accurately than single atlas-based models. For this purpose, majority voting and weighted sum methods were applied. The experimental findings clearly indicate that the multi-atlas model with a majority voting ensemble offers superior performance compared to the single-atlas model.  Our proposed model achieved improvements in accuracy between 4.08\% and 7.58\%, sensitivity between 2.59\% and 8.39\%, precision between 1.71\% and 8.42\%, and F1-score between 3.27\% and 6.77\% over other single-atlas models.

\bibliographystyle{agsm}
\bibliography{main}  






\end{document}